\documentclass[10pt,twocolumn,letterpaper]{article}

\usepackage{cvpr}
\usepackage{times}
\usepackage{epsfig}
\usepackage{graphicx}
\usepackage{amsmath}
\usepackage{amssymb}
\usepackage{subcaption}
\usepackage{algorithm,algpseudocode,mathrsfs,subcaption}
\usepackage{hhline}
\usepackage{multirow}

\usepackage[pagebackref=true,breaklinks=true,letterpaper=true,colorlinks,bookmarks=false]{hyperref}

\DeclareMathOperator*{\argmax}{arg\,max}
\DeclareMathOperator*{\argmin}{arg\,min}

\cvprfinalcopy 

\def\cvprPaperID{3387} 

\ifcvprfinal\fi
\begin{document}

\title{SuperMix: Supervising the Mixing Data
Augmentation}

\author{Ali Dabouei, Sobhan Soleymani, Fariborz Taherkhani, Nasser M. Nasrabadi\\
West Virginia University\\
{\tt\small \{ad0046, ssoleyma, ft0009\}@mix.wvu.edu,}
{\tt\small nasser.nasrabadi@mail.wvu.edu }}


\pagenumbering{gobble}

\maketitle

\begin{abstract}
This paper presents a supervised mixing augmentation method termed SuperMix, which exploits the salient regions within input images to construct mixed training samples. SuperMix is designed to obtain mixed images rich in visual features and complying with realistic image priors. To enhance the efficiency of the algorithm, we develop a variant of the Newton iterative method, $65\times$ faster than gradient descent on this problem. We validate the effectiveness of SuperMix through extensive evaluations and ablation studies on two tasks of object classification and knowledge distillation. On the classification task, SuperMix provides comparable performance to the advanced augmentation methods, such as AutoAugment and RandAugment. In particular, combining SuperMix with RandAugment achieves 78.2\% top-1 accuracy on ImageNet with ResNet50. On the distillation task, solely classifying images mixed using the teacher's knowledge achieves comparable performance to the state-of-the-art distillation methods. Furthermore, on average, incorporating mixed images into the distillation objective improves the performance by 3.4\% and 3.1\% on CIFAR-100 and ImageNet, respectively. {\it The code is available at https://github.com/alldbi/SuperMix}.
\end{abstract}


\section{Introduction}
Despite the revolutionary performance of deep neural networks (DNNs), they easily overfit when the training set is qualitatively or quantitatively deficient  \cite{srivastava2014dropout,zhang2016understanding}. Quality of the data can be interpreted as how well the data is expressive of the true distribution of inputs in the underlying task. This helps the model to learn discriminative patterns likely to occur at inference time. Quantity of the data, on the other hand, allows the model to observe discriminative patterns from different views and generalize the task-specific notions according to the major factors of variation in the input domain. Although analytical analysis of such important properties of the data has remained arduous \cite{kang2014convolutional}, empirical evaluations on training deep models often highlight a common observation: incorporating more data leads to a better generalization \cite{schmidt2018adversarially,hestness2017deep}. 
Hence, data augmentation has become a fundamental component of the training paradigms, aiming to enlarge the training set by transforming images in the given dataset. 

Conventional image data augmentation involves combinations of context-preserving transformations, such as horizontal flip, crop, scale, color manipulation, and cut out \cite{krizhevsky2009learning,han2017deep,devries2017improved}. Recently, notable efforts have been devoted to improving the augmentation, \eg, by automating the search for the optimal augmentation policies \cite{cubuk2019autoaugment,lim2019fast,cubuk2019randaugment}. The majority of these methods have focused on transforming single images, while ignoring the potentially very useful combination of multiple images for augmentation.
\begin{figure}[t]
     \centering
    \includegraphics[width=0.47\textwidth]{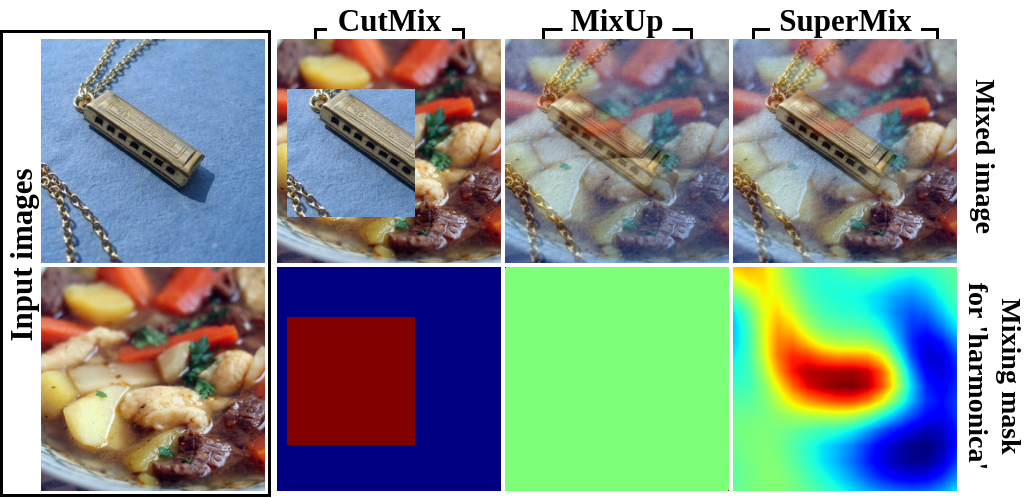}
     \small
    \caption{SuperMix combines salient regions in input images to construct unseen data for training.\vspace{-10pt}} 
    \label{fig:augment}
\end{figure}
To address this shortcoming, several studies have considered combining multiple images to construct novel images \cite{lemley2017smart,perez2017effectiveness, zhang2017mixup,yun2019cutmix,tokozume2018between}. However, these methods either mix images blindly and disregard the salient regions \cite{zhang2017mixup,guo2019mixup,yun2019cutmix,tokozume2018between} or do not scale to large-scale problems \cite{lemley2017smart}.  Furthermore, the current mixing functions are not expressive enough and often suppresses visual patterns by averaging or covering features in one image with the trivial features in another image. The corresponding pseudo labels are also not accurate and constrain the training performance \cite{guo2019mixup}. 

This paper presents a mixing augmentation approach termed SuperMix, which exploits the salient regions of input images to construct more advantageous mixed data. The supervision for this purpose can be obtained from the target model itself, \ie, self-training \cite{scudder1965probability,vapnik1998statistical,rosenberg2005semi,li2010optimol,chen2013neil,yarowsky1995unsupervised}, or a more sophisticated model aiming to guide a student network via knowledge transfer \cite{bucilu2006model,hinton2015distilling}. Figure \ref{fig:augment} provides a visual comparison of mixed images produced by different methods.  In a nutshell, the contributions of the paper are as follows:
\begin{itemize}
    \item We formalize the problem of supervised mixing augmentation using a set of mixing masks associating the pixel value at each spatial location in the mixed image to the spatial locations in the input images.
    \item  The optimization problem is carefully constrained to assure that the solutions are rich in salient features and comply with the realistic image priors.
    \item We develop a modified Newton iterative algorithm for SuperMix, suitable for large-scale applications. This approach provides $65\times$ speed-up as compared to SGD on ImageNet.
    \item 
    We demonstrate that mixed images intrinsically induce smooth predictions, and thus, help reveal knowledge of the teacher model in knowledge distillation. 
\end{itemize}

\section{Related work} \label{sec:relatedwork}
\noindent{\bf Data augmentation:} Data augmentation aims to improve the generalization of the model by enlarging the train set using transformations preserving the context of inputs in the learning problem. Conventional image transformations for this purpose are horizontal flip, crop, scale, color manipulation, and cut out \cite{krizhevsky2009learning,han2017deep,devries2017improved}. A contemporary trend of research on the topic has focused on selecting the best sequence of transformations according to the task, dataset, and learning model. AutoAugment (AA) \cite{cubuk2019autoaugment} automated the search for augmentation policies given a predefined set of transformations. Despite the significant performance of AA, it suffers from prohibitive training complexity imposed by Reinforcement Learning. Multiple approaches have attempted to reduce the training complexity by employing more efficient search methods, \eg, density matching in fast AutoAugment (FAA) \cite{lim2019fast}, or population based augmentation (PBA) \cite{ho2019population}. RandAugment (RA) \cite{cubuk2019randaugment} have shown that the search space and selection criteria can be significantly simplified by carefully combining random transformations. However, these methods ignore the potentially useful combination of multiple images for augmentation. 

\noindent{\bf Mixing augmentation:}
Several recent studies have considered employing multiple images for data augmentation \cite{lemley2017smart,zhang2017mixup,guo2019mixup,yun2019cutmix,tokozume2018between}. 
Smart Augmentation \cite{lemley2017smart} proposed merging multiple images from the same class using a DNN trained concurrently with the target model. However, training an additional deep model alongside every target model is resource exhaustive and severely limits the scalability of the approach for large-scale problems. Moreover, the method is restricted to merge images from the same class which limits the diversity and novelty of visual patterns in the merged images. MixUp \cite{zhang2017mixup,tokozume2018between} combined a pair of images for the augmentation by convex linear interpolation. CutMix \cite{yun2019cutmix} proposed overlaying a cropped area of an input image on another image to augment the data. Although MixUp and CutMix have demonstrated notable improvements to the training of object recognition models, they suffer from major shortcomings. First, they often average or replace salient regions in one image with insignificant regions, \eg, background, in another image. Second, due to the lack of supervision the labels computed for the mixed images are not accurate and limits the usefulness of the mixed images. However, SuperMix addresses these issues by extracting the salient regions of inputs and carefully combining them according to the realistic image priors and saliency-preserving constrains.    

\section{Supervised Mixing Augmentation}\label{sec:dataaugmentationfordistillation}
Given a training set $\mathcal{D}=\{(x_i, y_i)\}_{i=0}^{N-1}$, mixing methods take a subset $X\subset \mathcal{D}$ to produce the mixed image $\hat{x}$ and the corresponding label $\hat{y}$. A crucial property of mixed images is that they must reside close to the manifold of the training data since the goal of the mixing is to enlarge the support of the training distribution. Previous mixing methods \cite{zhang2017mixup,tokozume2018between} have considered this requirement by employing operations that preserve local smoothness of images.  MixUp \cite{zhang2017mixup,tokozume2018between} combines a pair of images $(x_i, x_j)$ using convex linear interpolation as: $\hat{x} = r x_i + (1-r) x_j$, where $r\sim \text{Beta}(\alpha, \alpha)$ is a random mixing weight from the symmetric Beta distribution with $\alpha \in (0,\infty)$. Due to the lack of supervision, the soft label for $\hat{x}$ is computed using the same linear interpolation as: $\hat{y} = r\delta(y_i) + (1-r)\delta(y_j)$, where $\delta(\cdot)$ is the one-hot encoding function. 
This blind mixing suffers from two shortcomings. First, coefficient $r$ assigns an equal importance to the whole 
image which can suppress important features by averaging with the background or less important features from the other image. Second, the computed soft label, $\hat{y}$, does not accurately describe the probability of classes represented by the mixed image and, thus, limits the effectiveness of the augmentation. 

\begin{figure*}[t]
    \centering
    \includegraphics[width=0.85\textwidth]{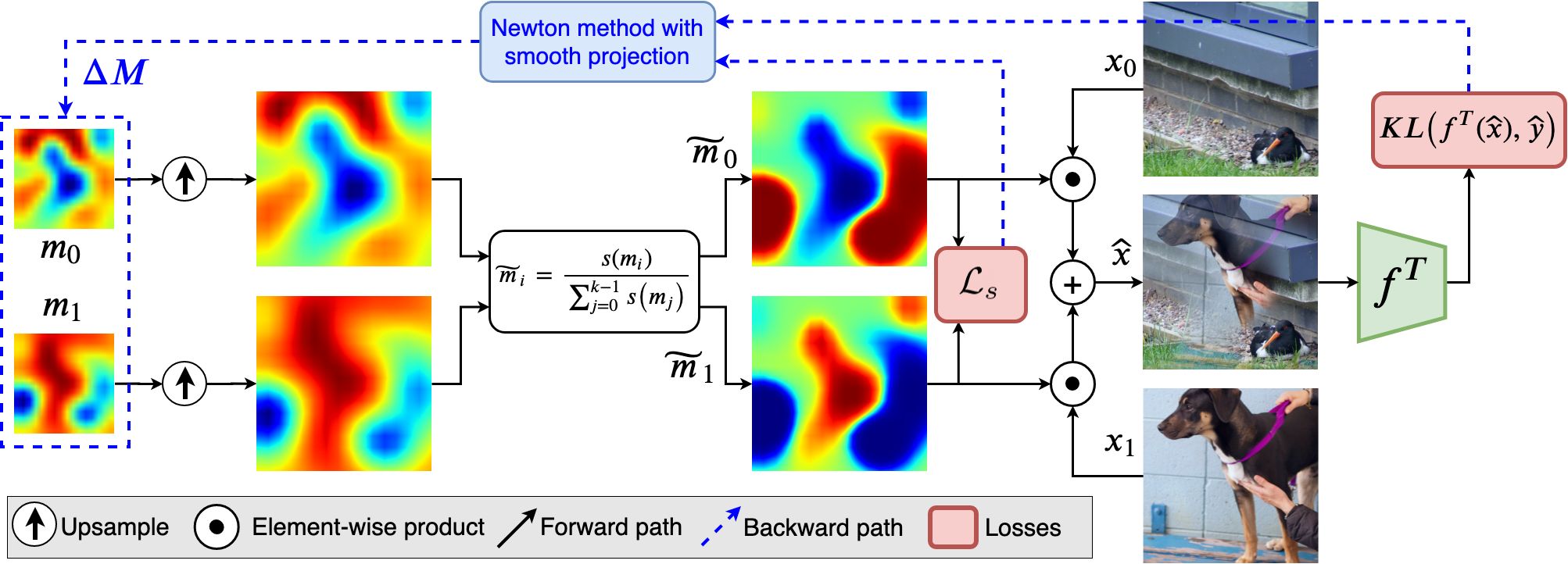}
    \caption{Schematic diagram of the proposed method for mixing $k=2$ input images using the supervision from $f^T$.}
    \label{fig:diagram}
\end{figure*}

\subsection{Mixing function}
We formalize a general formulation for the augmentation function that allows multiple images to be combined locally.  
We use a set of mixing masks $M=\{m_i\}_{i=0}^{k-1}$, where  $m_i:\Lambda \rightarrow [0, 1]$ associates each spatial location $u\in \Lambda$ in $x_i$ with a scalar value $m_i(u)$. Using the mixing masks, we define the mixing function as:
\begin{equation}\label{eq:mixaug}
    \hat{x}:= \sum\limits_{i=0}^{k-1} x_i \odot m_i,
\end{equation} 
where $x_i$ is the $i^{th}$ sample in $X$, the operator $\odot$ denotes the element-wise product, and $\sum_i m_i(u) = 1$ to hold the convexity of the combination.
The mixing function recovers MixUp \cite{zhang2017mixup} when $k=2$ and all values in each mask are equal. It also recovers CutMix \cite{yun2019cutmix} when $k=2$ and all values except the cropped area in one of the masks are equal to one. Figure \ref{fig:augment} provides a visual comparison of the role of the masks in the mixing augmentation. In the next section, we describe how knowledge of a teacher model can be used to compute $M$ such that the mixed image, $\hat{x}$, encompasses the rich visual information of images in $X$.

\subsection{Supervised mixing}\label{sec:supervisedmixing}

Let $f^T: \mathbb{R}^{W\times H \times C} \rightarrow [0, 1]^n$ denote the probability vector predicted by the teacher (T) for $n$ classes and $f_i^T$ be the probability for the $i^{th}$ class. We optimize the set of masks $M$ in Equation \ref{eq:mixaug} such that all salient regions in $X$, according to the knowledge of the teacher, be present in the mixed image, $\hat{x}$. 
This can be interpreted as: $f^T(\hat{x})\approx \hat{y}$, where $\hat{y}$ is high for classes associated with images in $X$. We formulate the target soft label, $\hat{y}$, computed in previous approaches \cite{zhang2017mixup,yun2019cutmix} for $k=2$ using the Beta distribution. We generalize for $k\geq 2$ by sampling the mixing coefficients from the Dirichlet distribution. Let $(r_0, \dots, r_{k-1})\sim \text{Dir}(\alpha)$ be a random sample from the symmetric multivariate Dirichlet distribution with parameter $\alpha$ and size $k$, we define the target soft label as:
\begin{equation}\label{eq:targetprob}
    \hat{y} := \sum_{i=0}^{k-1} r_i
    \delta\big(y^T(x_i)\big),
\end{equation}
where $y^T(x_i)=\argmax_j f^T_j(x_i)$ is the predicted class for $x_i \in X$, and $\delta(\cdot)$ is the one-hot encoding function. 

The set of mixing masks can be optimized to minimize the divergence between the output of the teacher model on the mixed image and the target soft label computed in Equation \ref{eq:targetprob}.   
The masks must also hold two additional properties to comply with the realistic image priors. First, generated images must reside close the manifold of the training data. In practice, this interprets that each mask must be spatially smooth so that the generated images resemble the spatial structure of the inputs. Second, masks must be sparse across the input samples to ensure each spatial location in the output image is assigned merely to a single image which prevents averaging multiple images at each spatial location and suppressing important features. Considering these, the optimization problem for finding the mixing masks can be written as:
\begin{equation}\label{eq:optimization1}
\begin{split}
    \argmin_{m_0,\dots,m_{k-1}} &KL(f^T(\hat{x})||\hat{y}) + \lambda_\sigma \mathcal{L}_\sigma(M) +  \lambda_s \mathcal{L}_s(M) ~~\text{s.t.:} \\ &
    a. ~ 0\leq m_i(u)\leq1, ~~b. ~ \sum\nolimits_i m_i(u)=1,
\end{split}
\end{equation}
where $\mathcal{L}_\sigma$ is a penalty term for the roughness of masks, \eg, total variation (TV) norm, $\mathcal{L}_s$ is a loss function to encourage sparsity of masks across input samples, and $KL(\cdot||\cdot)$ is the Kullback-Leibler divergence.

Here, we provide an iterative algorithm to solve the optimization problem efficiently. At each iteration $t$, the convexity conditions can be satisfied by the following normalization: 
\begin{equation}\label{eq:masknormalization}
    \widetilde{m}_i^t = \dfrac{s(m_i^t)}{\sum_{j=0}^{k-1} s(m_j^t)},
\end{equation}
where $s(\cdot)$ is the sigmoid function. Hence, the generalized mixing function in Equation \ref{eq:mixaug} takes the normalized masks to construct $\hat{x}$. Using the normalized masks, we define the sparsity promoting loss as:
\begin{equation}
    \mathcal{L}_s := \tfrac{1}{kWH} \sum_{u, i} \widetilde{m}_i^t(u)\big(\widetilde{m}_i^t(u)-1\big).
\end{equation}
This loss function encourages the mask values to approach $0$ or $1$. Since the values of masks at each spatial location sum to 1, due to the normalization in Equation \ref{eq:masknormalization}, only one of the masks takes the high value to minimize the loss.  

\begin{figure}[t]
    \centering
    \includegraphics[width=.47\textwidth]{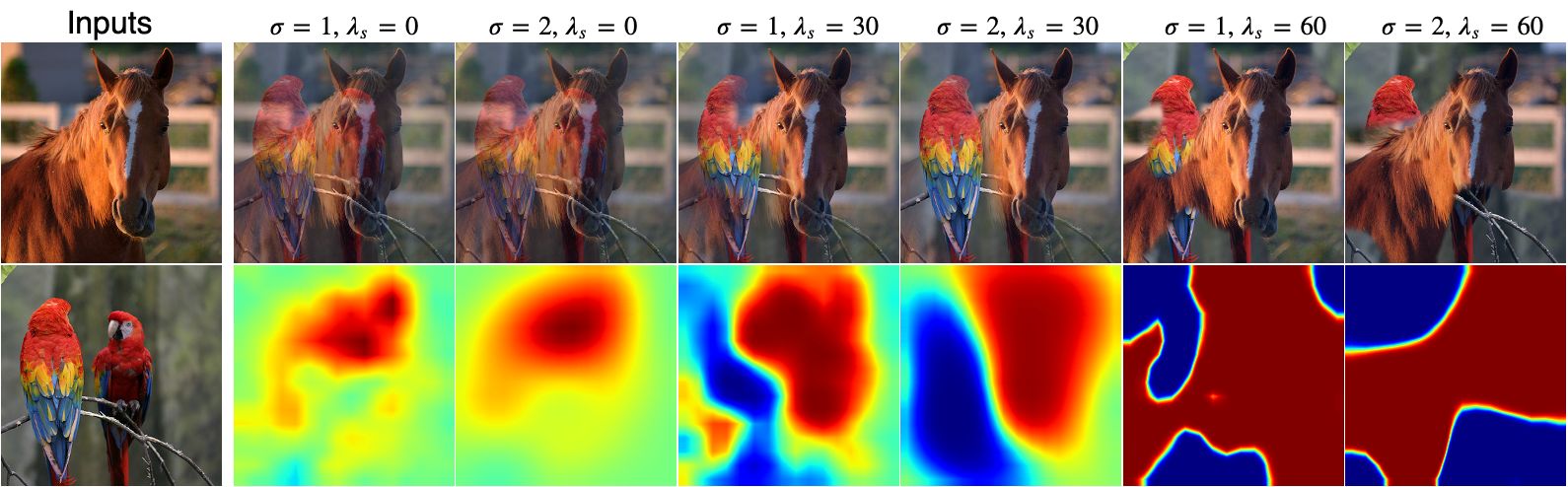}
    \caption{Visualizing the effect of smoothing factor, $\sigma$, and the sparsity promoting weight, $\lambda_s$, on the mixed images. Masks are estimated using ResNet34 and are associated with the `horse' class. }
    \label{fig:visualizingsmoothingandsparsity}
\end{figure}

\subsection{Optimization Method}
A proper set of mixing masks can be estimated by minimizing the objective of SuperMix as $\mathcal{L}_{SM}= KL+\lambda_\sigma\mathcal{L}_\sigma + \lambda_s\mathcal{L}_s$. A reduced form of this problem has been studied in saliency detection and explanation of DNN predictions by employing SGD \cite{fong2017interpretable} or deep generators \cite{dabkowski2017real}.
However, the current problem is more complex since multiple images are involved in the optimization and the roughness penalty and sparsity promoting loss should be minimized on all the corresponding masks. As we discussed and evaluated in Section \ref{sec:executiontime}, SGD is very slow and not feasible for solving the problem in case of large-scale image recognition tasks. Furthermore, employing a dedicated deep model to mix data by extending \cite{dabkowski2017real} 
makes the algorithm model-dependent and is not computationally efficient.

We develop a fast and efficient algorithm to optimize the mixing masks based on Newton’s iterative method for finding roots of a nonlinear system of equations in the underdetermined case \cite{moosavi2016deepfool,ruszczynski2006nonlinear}. Specifically, instead of optimizing $\mathcal{L}_{SM}$, we optimize $\mathcal{L}'_{SM}=KL+\lambda_s\mathcal{L}_s$ using a smooth projection (SP) \cite{dabouei2019smoothfool} that directly satisfies the smoothness of masks. As we analyze later in Section \ref{sec:executiontime}, this significantly improves the execution time of the mixing. Considering the first-order approximation of $\mathcal{L}'_{SM}$ at $M$, each mask is updated at iteration $t$ to find the roots as: $m_i^{t+1} \gets m_i^{t} + \Delta m^t_i$. Here, the update is computed using the Newton's method as: 
\begin{equation}\label{eq:newton}
    \Delta M^t = \dfrac{-|\mathcal{L}'_{SM}|}{||\nabla \mathcal{L}'_{SM} ||_2^2} \nabla \mathcal{L}'_{SM},
\end{equation}
where the gradient is with respect to $M^t$, the concatenation of $\{m_0^t,\dots, m_{k-1}^t\}$. Since both the divergence and $\mathcal{L}_s$ are nonnegative, $|\mathcal{L}'_{SM}|=\mathcal{L}'_{SM}$.
This formulation uses the $\ell_2$-norm projection to compute $\Delta M^t$. We modify it using SP to preserve the smoothness of masks and compute the smooth update as:
\begin{equation}\label{eq:smoothnewton}
    \widetilde{\Delta M}^t = \dfrac{-\mathcal{L}'_{SM}}{(g_\sigma * \nabla \mathcal{L}'_{SM})^T\nabla \mathcal{L}'_{SM}} (g_\sigma *\nabla \mathcal{L}'_{SM}),
\end{equation}
where $g_\sigma *\nabla \mathcal{L}'_{SM}$ is a smoothed version of the gradients using the 2D Gaussian smoothing filter $g$ with the standard deviation $\sigma$. It must be noted that all matrices in Equations \ref{eq:newton} and \ref{eq:smoothnewton} are vectorized before the matrix operations, and are reshaped back at the end of the iteration. In addition, due to the smoothness of masks, we optimize a down-sampled set of masks and up-sample them before performing the mixing. 
Algorithm \ref{alg:SMA} and Figure \ref{fig:diagram} demonstrate the detailed algorithm and schematic diagram for SuperMix, respectively.

\begin{algorithm}[t]
\small
\caption{SuperMix}
\label{alg:SMA}
\begin{algorithmic}[1]
\State \textbf{inputs:} Classifier $f^T$, set of $k$ images $X$,

low-pass filter $g_\sigma$.
\State \textbf{output:} Mixed sample $\hat{x}$.
\State $Y = \{\text{argmax}_j f^T_j(x_i): x_i \in X\}$.
\State Sample $(r_0, \dots, r_{k-1})$ from $\text{Dir}(\alpha)$.
\State $\hat{y} = \sum_{i=0}^{k-1} r_i
    \delta(y^T(x_i))$.
\State Initialize $(m_0,\dots$, $m_{k-1}) \gets 0$,

$\hat{x}^0 \gets \tfrac{1}{k}\sum_{x_i \in X} x_i  $, $t \gets 0$.
\State condition = Top-$k$ predicted classes by $f(\hat{x}^t)$ are not in $Y$.

\While{condition}
\State $\mathcal{L}'_{SM}=KL(f^T(\hat{x}^t)|| \hat{y})+\lambda_s\mathcal{L}_s$.
\State $\widetilde{\Delta} M^t = \tfrac{-\mathcal{L}'_{SM}}{(g_\sigma * \nabla \mathcal{L}'_{SM})^T\nabla \mathcal{L}'_{SM}} g_\sigma *\nabla \mathcal{L}'_{SM}$.
\State $m^{t+1}_i \gets m^t_i + \widetilde{\Delta}m_i$ for $i\in \{0,\dots,k-1\}$.
\State $\widetilde{m}_i^{t+1}=s(m_i^{t+1})/\sum_{j=0}^{k-1} s(m_j^{t+1})$.
\State $\hat{x}^{t+1} \gets \sum\limits_{i=0}^{k-1} x_i \odot \widetilde{m}^{t+1}_i$.
\State $t \gets t+1$
\EndWhile
\State \textbf{return} $\hat{x}^t$.
\end{algorithmic}
\end{algorithm}

\begin{figure*}[t]
     \centering
         \centering
         \includegraphics[width=0.8\textwidth]{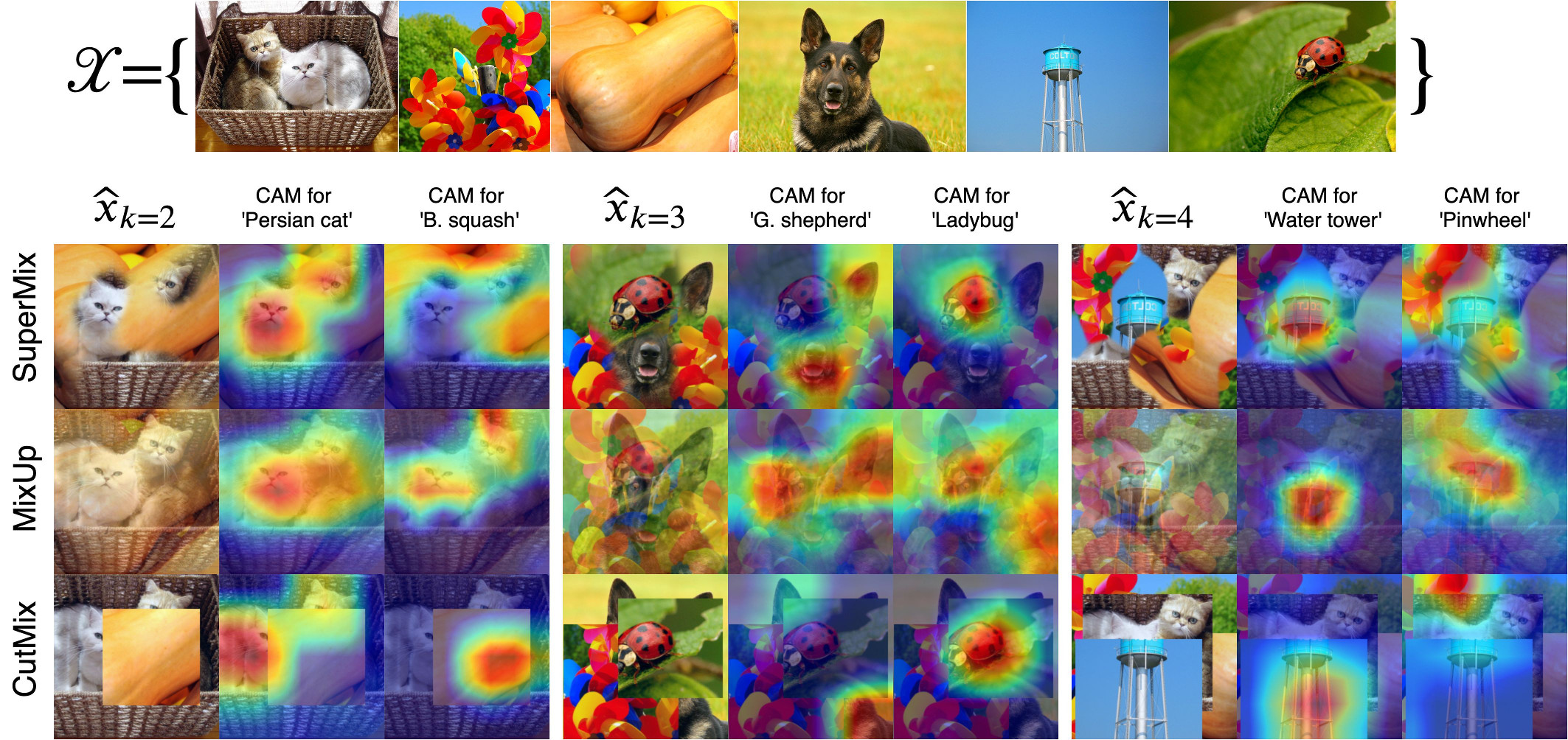}
         \caption{Visual comparison of the mixed images generated by SuperMix, MixUp, and CutMix, with $k\in \{2, 3, 4\}$ on ResNet34. Class activation maps \cite{zhou2016learning} are computed for two classes in mixed images. }
         \label{fig:visualcomparisonofk}
\end{figure*}

\noindent{\bf Termination Criteria:} The algorithm terminates when the Top-$k$ predicted classes of $f^T(\hat{x})$ are the same as the predicted class for samples in $X$. For instance, when $X$ consists of two images recognized as `cat' and `dog', the Top-$2$ classes in $f^T(\hat{x})$ should be classes of `cat' and `dog'. This criterion assures that important features in the input set are visible in the mixed image. Figure \ref{fig:visualcomparisonofk} provides a visual comparison of the mixed images produced by different methods.

\begin{table*}[t]
\centering
\small
\setlength{\tabcolsep}{0.4em}
\begin{tabular}{ll|c|ccc|ccc||c}\hline
\multirow{2}{*}{Dataset} & \multirow{2}{*}{Model} & \multirow{2}{*}{Base.}  & \multicolumn{3}{c}{Automated aug.} & \multicolumn{3}{|c||}{Mixing aug.} & SuperMix \\ \cline{4-9} 
 &   &  & AA\cite{cubuk2019autoaugment}  & FAA\cite{lim2019fast} & RA\cite{cubuk2019randaugment} & MixUp & CutMix & SuperMix & + RA\cite{cubuk2019randaugment} \\ \hline
\multirow{3}{*}{\begin{tabular}{c}
    CIFAR-  \\100
\end{tabular}} & WRN-40-2$_a$ & $74.0$  & $79.3$ & $79.4$ & $79.2$ & $77.2$ & $77.9$ & $\boldsymbol{79.7}$ & $79.9$ \\
 & WRN-28-10 & $81.2$ &  $82.9$ & $82.7$ & $83.3$ & $82.1$ & $82.9$ & $\boldsymbol{83.6}$ & $83.9$ \\ & S-S(26~2$\!\times\!$96d) & $82.9$ &  $\boldsymbol{85.7}$& $85.4$ & $85.6$ & $84.8$ & $85.0$  & $85.5$ & $85.8$ \\ \hline
\multirow{2}{*}{\begin{tabular}{c}
    ImageNet  \\
\end{tabular}} & ResNet-50 & $76.3/93.1$ &  $\boldsymbol{77.6/93.8}$  & $\boldsymbol{77.6}/93.7$ & $\boldsymbol{77.6/93.8}$ & $77.0/93.4$ & $77.2/93.5$ & $\boldsymbol{77.6}/93.7$ & $78.2/94.0$ \\
 & ResNet-200 & $78.5/94.2$ &  $80.0/95.0$  & $80.6/95.3$ & $80.4/95.3$ & $79.6/94.8$ & $79.9/94.9$ & $\boldsymbol{80.8/95.4}$ & $81.3/95.6$ \\\hline
\end{tabular}
\caption{Performance of augmentation methods on CIFAR-100 (Top-1 accuracy) and ImageNet (Top-1/Top-5 accuracy). \vspace{-10pt}}
\label{tab:selftraining}
\end{table*}

\begin{table*}[t]
    \centering
    \small
    \begin{tabular}{c|lccccccc|}
     \cline{2-9}
        &Teacher & \multicolumn{2}{c}{WRN-40-2$_b$} & ResNet56 & \multicolumn{2}{c}{ResNet110} & ResNet32x4 & VGG13 \\
        &Student & WRN-16-2 & WRN-40-1 & ResNet20 & ResNet20 & ResNet32 & ResNet8x4 & VGG8
        \\\cline{2-9}
        &Teacher acc. & \multicolumn{2}{c}{$75.61$} & $72.34$ & \multicolumn{2}{c}{$74.31$} & $79.42$ & $74.64$ 
        \\
        &Student acc. & $73.26$ & $71.98$ & $69.06$ & $69.06$ & $71.14$ & $72.50$ & $70.36$
        \\\hhline{-========}
        \multicolumn{1}{|l|}{\multirow{10}{*}{\rotatebox{90}{Distillation method}}}&KD \cite{hinton2015distilling} & $74.92$ & $73.54$ & $70.66$ & $70.67$ & $73.08$ & $73.33$ & $72.98$ \\
       \multicolumn{1}{|l|}{} &CRD \cite{tian2019contrastive} & $75.48$	& $74.14$ &	$71.16$ & $71.46$ &	$73.48$ &	$75.51$ &	$73.94$ \\\cline{2-9}
        
\multicolumn{1}{|l|}{}&\multirow{4}{*}{CE+}  ImgNet32 & $74.91$	& $74.80$ &	$71.38$ & $71.48$ &	$73.17$ &	$75.57$ &	$73.95$ \\
\multicolumn{1}{|l|}{}&~~~~~~~~~MixUp & ~$76.20^\star$	& $75.53$ &	$72.00$ & $72.27$ &	~$74.60^\star$ &	$76.73$ &	$74.56$ \\
\multicolumn{1}{|l|}{}&~~~~~~~~~CutMix & ~$76.40^\star$ & ~$75.85^\star$ &	$72.33$ & $72.68$ &	$74.24$ &	$76.81$ &	~$74.87^\star$ \\
\multicolumn{1}{|l|}{}&~~~~~~~~~SuperMix & ~~$\boldsymbol{76.93}^\star$	& ~~$\boldsymbol{76.11^\star}$ &	~~$\boldsymbol{72.64^\star}$ & $\boldsymbol{72.75}$ &	~~$\boldsymbol{74.80^\star}$ &	$\boldsymbol{77.16}$ &	~~$\boldsymbol{75.38^\star}$ \\    \cline{2-9}

\multicolumn{1}{|l|}{}&\multirow{4}{*}{KD+} ImgNet32 & ~$76.52^\star$	& ~$75.70^\star$ &	$72.22$ & $72.23$ &	$74.24$ &	$76.46$ &	~$75.02^\star$\\
\multicolumn{1}{|l|}{}&~~~~~~~~~MixUp & ~$76.58^\star$	& ~$76.10^\star$ &	~$72.89^\star$ & $72.82$ &	~$74.94^\star$ &	$77.07$ &	~$75.58^\star$ \\
\multicolumn{1}{|l|}{}&~~~~~~~~~CutMix & ~$76.81^\star$& ~$76.45^\star$ &~$72.67^\star$ & $72.83$ & ~$74.87^\star$ &	$76.90$ &	~$75.50^\star$ \\    
\multicolumn{1}{|l|}{}&~~~~~~~~~SuperMix & $~~\boldsymbol{77.45^\star}$	& $~~\boldsymbol{76.53^\star}$ &	$~~\boldsymbol{73.19^\star}$ & $\boldsymbol{72.96}$ &	$~~\boldsymbol{75.21^\star}$ &	$\boldsymbol{77.59}$ &	$~~\boldsymbol{76.03^\star}$ 
 \\\hline

    \end{tabular}
    \caption{Classification performance (\%) of student models on CIFAR-100. Teacher and student are from the same architecture family but different depth/wideness and capacity. We denote by $\star$ results where the student surpasses the teacher performance. Only ImgNet32 uses unlabeled data from an external source.  Average over 4 independent runs.}
    \label{tab:resultssamenets}

\end{table*}

 \begin{figure*}[t]
     \centering
         \includegraphics[width=0.96\textwidth]{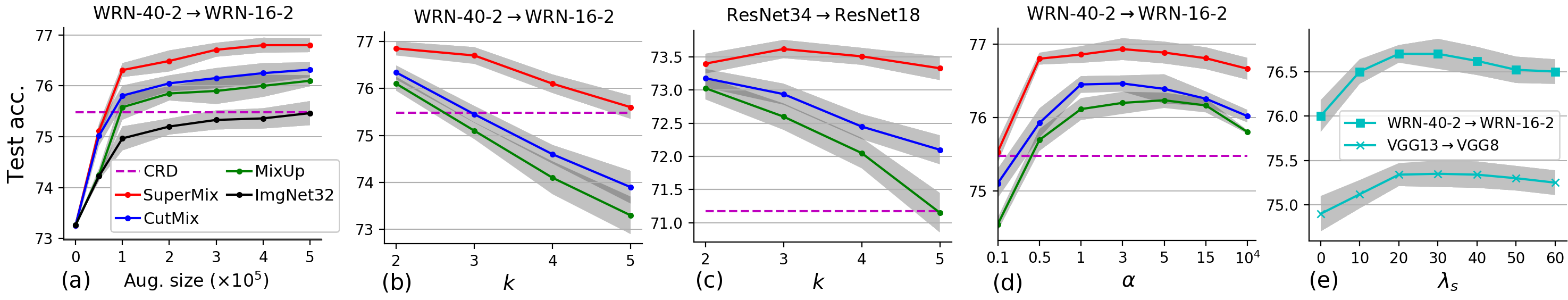}
        \caption{Evaluating the role of augmentation size and hyper-parameters.}
        \label{fig:ablation}
\end{figure*}

\section{Experiments}
We evaluate the performance of SuperMix on two tasks of object classification and knowledge distillation \cite{bucilu2006model,hinton2015distilling} 
using two benchmark datasets of CIFAR-100 \cite{krizhevsky2009learning} and ImageNet \cite{deng2009imagenet}. For knowledge distillation, we evaluate SuperMix on two major previous SOTA methods \cite{hinton2015distilling,tian2019contrastive} and two mixing augmentation techniques including MixUp and CutMix. For the sake of fair comparison, {\it pseudo labels for these blind mixing methods are computed using the same teacher employed in SuperMix}. All training experiments use random horizontal flip and random crop as the default augmentations. We perform the algorithm on random sets of input samples drawn from $\mathcal{D}$ to generate $\mathcal{D}'$. For the sake of brevity, we define the augmentation factor $\kappa=\tfrac{|\mathcal{D}'|}{|\mathcal{D}|}$ to show the ratio of the size of the mixed dataset over the size of the original dataset.

\vspace{-10pt}
For knowledge distillation on CIFAR-100, we also consider an additional baseline by using unlabeled data from 
the training set of ImageNet32x32 \cite{chrabaszcz2017downsampled} (ImgNet32) to construct unlabeled sets. This helps to better evaluate the role of the data provided by the mixing augmentation methods. We use SGD optimizer with an initial learning rate of $0.1$ and momentum of $0.9$. Weight decay is set to $5e-4$. The learning rate is decayed by $0.1$ at epochs $200, 300, 400,$ and $500$, and the maximum number of epochs is set to $600$. Since in our experiments $\kappa\geq 1$, the number of epochs according to the mixed dataset will scale with $\tfrac{1}{\kappa}$ to keep the number of training iterations fixed for all experiments. For instance, when $\kappa=5$, the maximum number of epochs for the mixed dataset is $120$. The batch size is set to $128$ and $256$ for CIFAR-100 and ImageNet, respectively. For the CIFAR-100 dataset, we set $\sigma$ of the Gaussian smoothing in SuperMix to $1$ and the spatial size of the masks to $8\times8$. For ImageNet, $\sigma$ is set to $2$ and the size of masks is set to $16\times16$. For all benchmark comparisons, we set $\alpha=3$ and $\lambda_s=25$. Moreover, in all experiments, the performance of SuperMix is evaluated by generating $5\times10^5$ and $10^6$ images on CIFAR-100 and ImageNet, respectively, unless otherwise noted.   
All the hyper-parameters for the distillation experiments are selected according to the experimental setup of \cite{tian2019contrastive} and the ablation studies in Section \ref{sec:ablationstudies}. Network architectures and settings for baseline methods are provided in the supplemental material. 
\subsection{Object classification}
We follow the standard setup of evaluation for automated augmentation \cite{cubuk2019autoaugment,lim2019fast,ho2019population} and compare them with SuperMix
on the task of object classification. For SuperMix, we first train the target model on the original dataset and then use it to generate mixed data with $k$ equal to $2$ and $3$ for CIFAR-100 and ImageNet, respectively. Afterward, we train the target model from scratch on the mixture of the augmented data and the original data. Rest of the result are reported from the original papers. As an additional evaluation, we combine SuperMix with RangAugment (RA) \cite{cubuk2019randaugment}. For this purpose, we first mix images using SuperMix and then apply RA with the default parameters \cite{cubuk2019randaugment} for CIFAR-100 and ImageNet.  Table \ref{tab:selftraining} presents the results for these experiments. On four out of five experiments, SuperMix provide performance competitive to SOTA approaches of automated augmentation. Furthermore, combining RA with SuperMix further improves the performance of classification across all the experiments. These evaluations highlight the effectiveness of mixing multiple images for data augmentation.

\subsection{Knowledge Distillation}\label{sec:resultsoncifar100}
In addition to KD \cite{hinton2015distilling} and CRD \cite{tian2019contrastive}, we consider a simple method for distillation to highlight the effectiveness of mixing augmentation. In this method, we train the student models to classify mixed images labeled by the teacher model. The labels only show the winner class and does not contain any information regarding the rest of the classes. We refer to this method as Cross-Entropy (CE) distillation.   

\noindent{\bf Results on CIFAR-100:} 
Tables \ref{tab:resultssamenets} and \ref{tab:resultsdifferentnets} presents the results for two challenging scenarios of distillation. In the first scenario, teacher and student are from the same family of architectures but have different depth/wideness and capacity. In the second scenario, teacher and student are from completely different network architectures. Employing the simple CE method using the mixed data consistently outperforms previous methods in both distillation scenarios. The data generated by SuperMix demonstrates the best performance across all evaluations, and, on five out of seven teacher-student setups from the same architecture family, students trained on the SuperMix data outperform their teachers.  
Last four rows in Tables \ref{tab:resultssamenets} and \ref{tab:resultsdifferentnets} present the results for knowledge distillation using the original KD \cite{hinton2015distilling}. More importantly, results on MixUp, CutMix, and SuperMix demonstrate that they can notably enhance the performance of the distillation techniques.

These observations highlight three crucial points. First, the limited size of the training set is a major factor constraining the performance of knowledge distillation. According to Table \ref{tab:resultssamenets}, almost all of the students achieve comparable results to CRD when external data of ImgNet32 is provided. Second, mixing augmentation provides more informative data for distillation compared to unlabeled data from an external source. Third, the supervised mixing results in rich images that are highly favorable for  knowledge distillation and outperforms blind mixing methods.   
\begin{table*}[t]
    \centering
    \small
    \setlength{\tabcolsep}{0.3em}
    \begin{tabular}{c|lcccccc|}
     \cline{2-8}
        &Teacher & VGG13 & \multicolumn{2}{c}{ResNet50} & \multicolumn{2}{c}{ResNet32x4}  & WRN-40-2 \\
        &Student & MobileNetV2 & MobileNetV2 & VGG8 & ShuffleNetV1 & ShuffleNetV2 & ShuffleNetV1
        \\\cline{2-8}
        &Teacher acc. & $74.64$ & \multicolumn{2}{c}{$79.34$} & \multicolumn{2}{c}{$79.42$} & $75.61$  \\
        &Student acc. & $64.60$ & $64.60$ & $70.36$ & $70.50$ & $71.82$ & $70.50$ 
        \\
        \hhline{-=======}
        \multicolumn{1}{|l|}{\multirow{10}{*}{\rotatebox{90}{Distillation method}}}&KD \cite{hinton2015distilling} & $67.37$ & $67.35$ & $73.81$ & $74.07$ & $74.45$ & $74.83$  \\

        \multicolumn{1}{|l|}{}&CRD \cite{tian2019contrastive} & $69.73$& $69.11$ &	$74.30$ & $75.11$ &	$75.65$ &	~$76.05^\star$  \\
        \cline{2-8}
        \multicolumn{1}{|l|}{}&\multirow{4}{*}{CE+} ImgNet32 & $68.85$	& $68.01$  & $73.96$ & $76.80$ &	$77.56$ &	$~75.87^\star$  \\        
        \multicolumn{1}{|l|}{}&~~~~~~~~~MixUp & $71.13$	& $71.71$  &	$75.41$ & $78.16$ &	$78.84$ &	~$77.29^\star$  \\
        \multicolumn{1}{|l|}{}&~~~~~~~~~CutMix & $70.93$ & $70.64$  &	$75.84$ & $77.89$ &	$79.32$ &	~$77.50^\star$  \\
        \multicolumn{1}{|l|}{}&~~~~~~~~~SuperMix & $~\boldsymbol{71.65}$	& $~\boldsymbol{72.13}$ &	$~\boldsymbol{76.07}$ & $~\boldsymbol{78.47}$ &	$~~\boldsymbol{79.53^\star}$ &	$~~\boldsymbol{77.92^\star}$ \\ \cline{2-8}

        \multicolumn{1}{|l|}{}&\multirow{4}{*}{KD+} ImgNet32 & $69.14$ & $68.44$ & $74.32$ & $76.87$ &	$77.90$ &	$~76.23^\star$ \\ 
        \multicolumn{1}{|l|}{}&~~~~~~~~~MixUp  & $71.29$ & $71.99$ & $75.59$ & $78.22$ & $79.14$ & $~77.44^\star$  \\
        \multicolumn{1}{|l|}{}&~~~~~~~~~CutMix &  $71.10$ & $70.93$ & $76.01$ & $77.92$ & $~79.53^\star$ & $~77.65\star$ \\
        \multicolumn{1}{|l|}{}&~~~~~~~~~SuperMix  & $~\boldsymbol{71.81}$ & $~\boldsymbol{72.40}$ & $~\boldsymbol{76.28}$ & $~\boldsymbol{78.51}$ & $~~\boldsymbol{79.80^\star}$ & $~~\boldsymbol{78.07^\star}$  \\
        \hline    
    \end{tabular}
    \caption{Classification performance (\%) of student models on CIFAR-100. Teacher and student models are from different architectures. We denote by $\star$ results where the student surpasses the teacher performance. Average over 4 independent runs.   }
    \label{tab:resultsdifferentnets}

\end{table*}

\noindent{\bf Results on ImageNet:}
We showcase the effectiveness of the mixed data on ImageNet by distilling the knowledge of ResNet-34 into ResNet-18. Table \ref{tab:imgnetresults} presents the results for the distillation on the ImageNet dataset. Using the simple CE method consistently outperforms the previous SOTA approaches. In five out of eight experiments of distillation using mixed images, the student outperforms the teacher. This demonstrates the scalability and effectiveness of the mixing augmentation for the task of knowledge distillation. Moreover, combining mixed data with the original distillation objective further enhances the distillation performance validating the effectiveness of the mixing augmentation for knowledge transfer in large-scale datasets.   

\subsection{Ablation studies}\label{sec:ablationstudies}
\noindent{\bf Impact of the size of the training set:}
In this part, we investigate how the size of the dataset affects the distillation performance by measuring the Top-1 test accuracy of WRN-16-2 versus the augmentation size on CIFAR-100. For all the mixing methods, we set $k=2$ and $\alpha=1$, \ie, sampling mixing coefficients from the uniform distribution. Figures \ref{fig:ablation}a presents the results for these evaluations. The distillation performance improves by increasing the augmentation size and plateaus at $5\times10^5$. All the datasets generated using mixing augmentations outperform the unlabeled dataset of ImgNet32. This highlights the superiority of mixed images for knowledge transfer compared to unlabeled data from an external source.  
Based on these observations, we set the size of the mixed dataset to $5\times 10^5$ for all experiments on CIFAR-100.

\begin{table}
\centering
     \scriptsize
\begin{tabular}{lllccc}
\multirow{2}{*}{} & \multirow{2}{*}{Net} & & \multirow{2}{*}{SGD} & \multicolumn{2}{c}{Newton} \\ \cline{5-6} 
 & &  &  & w/o SP & w/ SP \\ \hline
\multirow{4}{*}{\rotatebox{90}{ImgNet}} & \multirow{2}{*}{VGG16} & $ET(sec.)$ & $15.41$ & $6.59$ &  $\boldsymbol{0.23}$ \\
 &  &  $iters$ & $34.5$ & $15.1$ & $\boldsymbol{0.5}$  \\ \cline{2-6} 
 & \multirow{2}{*}{Res34} & $ET(sec.)$ & $4.25$ & $1.98$ & $\boldsymbol{0.06}$   \\
 &  & $iters$ & $23.6$ & $11.7$ & $\boldsymbol{0.3}$   \\ \hline
\multirow{4}{*}{\rotatebox{90}{CIFAR}} & \multirow{2}{*}{VGG13} & $ET(ms.)$ & $482$ & $97$  & $\boldsymbol{5}$  \\
 &  & $iters$ & $19.5$ & $3.7$ & $\boldsymbol{0.2}$  \\
\cline{2-6} 
 & \multirow{2}{*}{WRN} & $ET(ms.)$ & $509$ & $122$ & $\boldsymbol{6}$  \\
 &  & $iters$ & $21.8$ & $4.6$ & $\boldsymbol{0.2}$   \\ \hline
\end{tabular}
        \caption{Comparison of execution time.}
        \label{tab:executiontime}
\end{table}

\noindent{\bf Impact of $\boldsymbol{k}$:} We evaluate the role of $k$ by conducting experiments on CIFAR-100 and ImageNet datasets. Figures \ref{fig:ablation}b and \ref{fig:ablation}c present the results for this evaluation. A major shortcoming of MixUp and CutMix is that they mix images without any supervision. Including more input images to produce a mixed image increases the chance of incorrect cropping in CutMix, and averaging overlapping features in Mixup. This explains the notable deterioration of the distillation performance in all experiments with $k>2$ using these augmentation methods.  Both of these incidents degrade the quality and effectiveness of features in the mixed image, which can also be observed from the visual comparisons provided in Figure \ref{fig:visualcomparisonofk}. We observe that the spatial size of the image can limit $k$. Performance of distillation using SuperMix degrades for $k>2$ on CIFAR-100. However on ImageNet, $k=3$ yields the best distillation performance. 

\noindent{\bf Impact of $\boldsymbol{\alpha}$:}
Parameter $\alpha$ determines the probability distribution for the presence of each input class in the mixed image. We measure the performance of distillation versus several values of $\alpha$ to identify its optimal value. Figure \ref{fig:ablation}d presents results for these experiments. For $\alpha \rightarrow 0$, the mixing augmentation becomes inactive since only one input category will appear in the augmented images, \ie, $r_0=1$ or $r_1=1$. For $\alpha \rightarrow +\infty$, the contribution of images become equal, \ie, $r_0=r_1=0.5$. This is more favorable for distillation since both input images contribute equally to the mixed image. For $\alpha=1$, contribution of each input in the mixed image is selected from the uniform distribution $\text{Unif}(0, 1)$. According to the figures, we select $\alpha=3$ for all other experiments unless otherwise noted.

\begin{table*}[t]
\centering
    \small
        \setlength{\tabcolsep}{0.7em}
\begin{tabular}{c|cc|cc|cccccccc} 
\cline{2-13}
& \multirow{2}{*}{\rotatebox{40}{Teacher}} & \multirow{2}{*}{\rotatebox{40}{Student}}
 & \multirow{2}{*}{KD} &  \multirow{2}{*}{CRD} &  CE & KD & CE & KD & CE & KD & CE & KD \\ 
     & & & & &  \multicolumn{2}{c}{+MixUp$_{k=2}$} & \multicolumn{2}{c}{+CutMix$_{k=2}$} & \multicolumn{2}{c}{+SuperMix$_{k=2}$} & \multicolumn{2}{c}{+SuperMix$_{k=3}$} \\   \hline
     Top-1 & $73.31$ & $69.75$ &   $70.66$  & $71.17$  & $73.03$ & $73.29$ & $73.18$ & $73.33^\star$ & $73.42^\star$ & $73.62^\star$ & $73.65^\star$ & $\boldsymbol{73.83^\star}$ \\
     Top-5 & $91.42$ & $89.07$ &  $89.88$  & $90.13$    & $91.27$ & $91.44$ & $91.36$ & $91.44^\star$ & $91.51^\star$ & $91.66^\star$ & $91.67^\star$ & $\boldsymbol{91.82^\star}$\\\hline
\end{tabular}
    \caption{Top-1 and Top-5 classification accuracy of ResNet18 on ImageNet dataset. 
    Results where the student surpasses the teacher performance are marked by $\star$. Average over 4 independent runs.}
    \label{tab:imgnetresults}
\end{table*}

\noindent{\bf Sparsity among masks:} The sparsity promoting loss forces each spatial location in the output image to be assigned to only one image in the input set. This improves the mixing performance by preserving the most important features in each spatial location. We evaluate the performance of distillation versus $\lambda_s$ in Figure \ref{fig:ablation}e. By increasing the weight of sparsity the performance of distillation improves until $\lambda_s\approx 30$. After that the accuracy of masks degrades since the sparsity promoting loss dominates the $KL$ loss. Figure \ref{fig:visualizingsmoothingandsparsity} evaluates this phenomenon by visualising the mixing mask versus $\lambda_s$.

\subsection{Execution time}\label{sec:executiontime}
Here, we compute the execution time of SuperMix. To this aim, we define two baselines for the sake of comparison. For the first baseline, we use SGD instead of the Newton method to optimize the set of masks. The second baseline is the Newton method without SP. Hence, the optimization in both baselines is performed on $\mathcal{L}_{SM}= KL+\lambda_\sigma\mathcal{L}_\sigma + \lambda_s\mathcal{L}_s$. Inspired by the previous work on saliency detection \cite{fong2017interpretable}, we use the TV norm for the spatial smoothness loss as:
$\mathcal{L}_s = \tfrac{1}{kWH}\sum_i\sum_{u\in \Lambda} ||\nabla m_i(u)||_3^3 $. Based on experimental observations, we set $\lambda_s=250$, learning rate of SGD to $0.1$. All other parameters are set to the values identified in previous sections. All algorithms are implemented with parallel processing on two NVIDA Titan RTX with batch size of 128. For further implementation details, please refer to the released code.

Figure \ref{tab:executiontime} presents the results for these comparisons. Newton method with SP, \ie, SuperMix, is at least $\boldsymbol{65\times}$ faster than SGD on both datasets. Moreover, due to SP which directly satisfied the spatial smoothness condition, SuperMix is at least $\boldsymbol{19\times}$ faster than the same algorithm when it has to include $\mathcal{L}_s$.

\begin{figure}
     \centering
         \centering
         \includegraphics[width=0.35\textwidth]{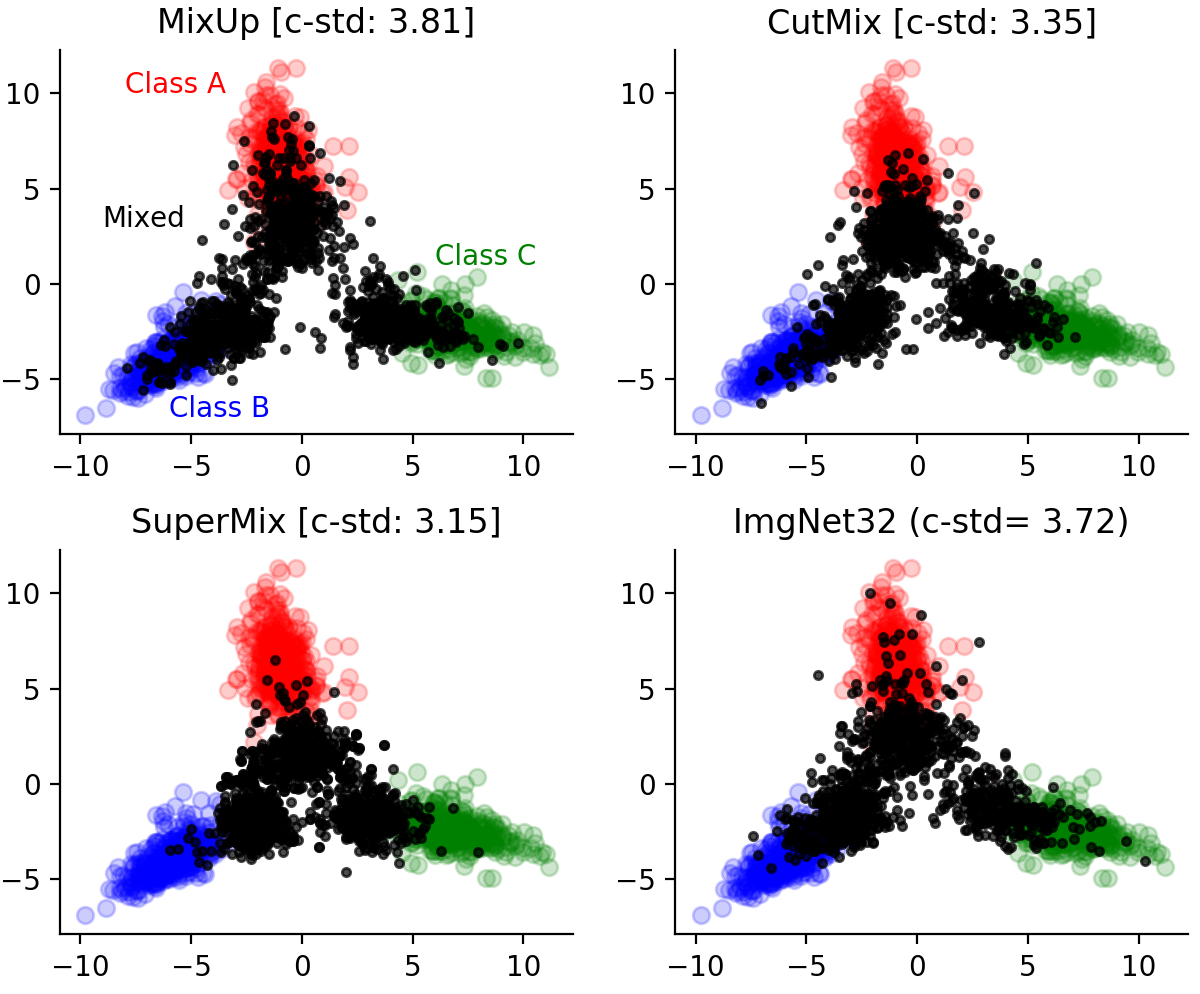}
         \caption{Visualizing representations for the mixed images. \vspace{-20pt} }
         \label{fig:visembeddings}
\end{figure}

\subsection{Embedding space evaluations}
We perform two sets of evaluations on CIFAR-100 to further analyze characteristics of the mixed images. In the first set of experiments, we feed the original data and the mixed images to VGG13 and visualize the output of the logits layer, in 2D for three random classes using PCA. The SuperMix images are generated with $k=2$. Figure \ref{fig:visembeddings} demonstrates these evaluations. Representations for the SuperMix data has less overlap with the distribution of the representations for the original data. This suggests that the SuperMix data encompass more novel structure compared to the original data, unlabeled data from other mixing methods or an external source. The SuperMix data are harder to classify for the model since the representations are concentrated close to the center of the embedding. To better evaluate this, we compute the class standard deviation (c-std) of representations for each class. The computed values are reported on the top of the corresponding images in Figure \ref{fig:visembeddings}. 

\begin{figure}
     \centering
     \includegraphics[width=0.33\textwidth]{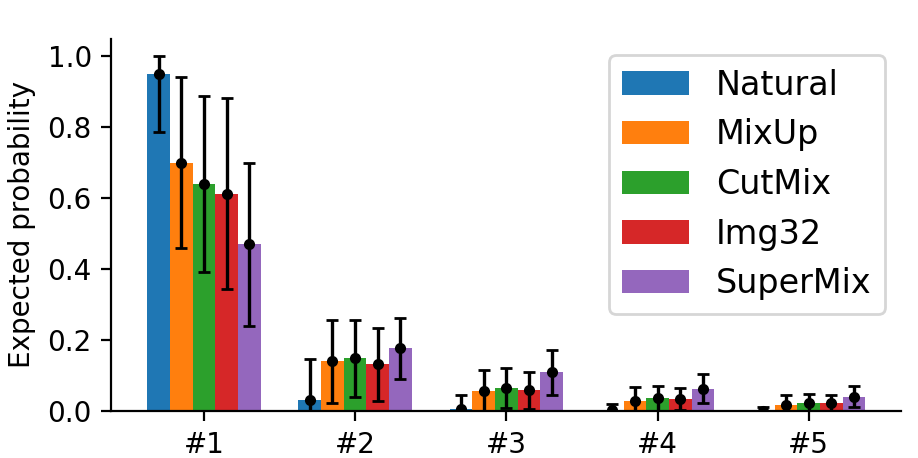}
     \caption{Distribution of top 5 predictions.\vspace{-10pt}}
     \label{fig:distribution}
\end{figure}

Hinton \etal \cite{hinton2015distilling} pointed that smoothing out the predictions of a model can better reveal its knowledge of the task. Since SuperMix generates images by combining multiple inputs, the outputs of the model on SuperMix data are intrinsically more smooth compared to that of the other augmentation types. We validate this by computing the average of the sorted Top-5 probability predictions of VGG13 on the original and augmented images of CIFAR-100. 
As demonstrated in Figure \ref{fig:distribution}, predictions of the target model is significantly smoother on mixed images. Moreover, SuperMix produces the data with the most smooth labels.  

\section{Conclusion}
In this paper, we studied the potential of mixing multiple images using supervision of a teacher for the data augmentation. We proposed SuperMix, a supervised mixing augmentation method that combines salient regions in multiple images to produce unseen training samples. The effectiveness and efficiency of SuperMix is validated through extensive experiments, evaluations, and ablation studies. Specifically, incorporating SuperMix data for distillation enhances the state of the art of knowledge distillation. SuperMix provides comparable performance to the automated augmentation methods, and when combined, notably improves the generalization of the model. 

{
\bibliographystyle{ieee_fullname}
\bibliography{arxiv.bib}
}

\end{document}